\documentclass[]{article}

\usepackage{authblk}
\usepackage[pdftex]{graphicx,xcolor}
\usepackage{newtxtext}
\usepackage{cite}
\usepackage[cmex10]{amsmath}
\usepackage{lmodern}
\usepackage{textcomp}
\usepackage{latexsym}
\usepackage{amsfonts}
\usepackage{bm}
\usepackage{subfig}
\usepackage{amsmath}
\usepackage{colortbl}

\graphicspath{{./figures/}}
\title{Anomaly Detection and Interpretation using Multimodal Autoencoder and Sparse Optimization}

\author[1]{Yasuhiro Ikeda}
\author[2]{Keisuke Ishibashi}
\author[1]{Yuusuke Nakano}
\author[1]{Keishiro Watanabe}
\author[3]{Ryoichi Kawahara}
\affil[1]{NTT Network Technology Laboratories, NTT Corporation}
\affil[2]{Division of Arts and Sciences, College of Liberal Arts, International Christian University}
\affil[3]{Faculty of Information Networking for Innovation and Design, Toyo University}
\affil[ ]{\textit{yasuhiro.ikeda.sm@hco.ntt.co.jp, ikeisuke@icu.ac.jp, \{yuusuke.nakano.dn,  keishiro.watanabe.ry\}@hco.ntt.co.jp, kawahara011@toyo.jp}}

\date{}

\begin{document}
\maketitle
\begin{abstract}
Automated anomaly detection is essential for managing information and communications technology (ICT) systems to maintain reliable services with minimum burden on operators.
For detecting varying and continually emerging anomalies as differences from normal states,
learning normal relationships inherent among cross-domain data monitored from ICT systems is essential.
Deep-learning-based anomaly detection using an autoencoder (AE) is therefore promising for such complicated learning;
however, its interpretation is still problematic.
Since the dimensions of the input data contributing to the detected anomaly are not directly indicated in an AE,
they are not suitable for localizing anomalies in large ICT systems composed of a huge amount of equipment.
We propose an algorithm using sparse optimization for estimating contributing dimensions to anomalies detected with AEs.
We also propose a multimodal AE (MAE) for effectively learning the relationships among cross-domain data, which can induce nonlinearity and differences in learnability among data types.
We evaluated our algorithms with several datasets including real measured data in comparison with conventional algorithms and confirmed the superiority of our estimation algorithm in specifying contributing dimensions of anomalous data
and our MAE in detecting anomalies in cross-domain data.
\footnote{This paper is an extended version of our technical report~\cite{ike2017} with no peer review. The difference is the proposal of our MAE and additional evaluations with real measured data.}
\end{abstract}

\section{Introduction}
Fault diagnosis and recovery are essential for managing information and communications technology (ICT) systems to maintain reliable services. Although the growth of ICT systems, which involves virtualization, large cloud servers, and so on, enables advanced and flexible services, it increases the types of system faults; thus, burdening operators. Anomalies thus need to be automatically detected and localized through surveillance of ICT systems as a part of fault diagnosis for reducing not only the operator's workload but also the effect on the services. For detecting varying and continually emerging anomalies of ICT systems, ``normal states'' of the systems need to be learned from monitored data
so that anomalies appear as differences from the normal states. In ICT systems, however, complicated relationships are inherent among cross-domain data, which are data from different domains, such as management information base (MIB) data, which represent network load; network flow data, which represent service usage; and syslog data, which represent states of equipment.
For example, as Govindan et al. discussed~\cite{gov2016}, lack of synchronization between the control and data planes can cause various types of failures in software-defined networks, and the consistency of invariants between these planes should be maintained. The invariants will appear as dependencies between cross-domain data, such as syslogs monitored from the control plane and MIB data monitored from the data plane.

Network management systems (NMSs) ~\cite{nms01, nms03} are often used for network surveillance on the basis of integrative analysis of cross-domain data.
However, the rule-based analysis adopted in these systems is not sufficient since manually determining adequate rules, such as threshold, for individual monitored metrics to express the complicated relationships among cross-domain data is impractical.
Automatically learning normal relationships from monitored data for detecting anomalies as differences from normal states is therefore essential.
Although several machine-learning approaches have been proposed~\cite{rin2007, nec, ibm} for anomaly detection,
such as using principal component analysis (PCA)~\cite{rin2007},
conventional approaches have assumed linear relationships among data.
The relationships among quantitative data, such as traffic volume and the number of packets, 
may often become linear.
However, cross-domain data that include syslog messages can introduce nonlinear relationships;
therefore, conventional linear approaches are not sufficient.

Deep-learning approaches using autoencoders (AEs)~\cite{tho2002, sak2014, ara2016, zho2017}
are therefore promising for learning the nonlinear relationships among cross-domain data.
Although applications of AEs to anomaly detection have been extensively discussed,
their interpretation is still problematic.
AEs can calculate the degree of abnormality of multidimensional input data as one metric
according to the difference from reconstructed data from the low-dimensional expression of the input data.
For surveillance of large ICT systems, however, the dimensions of input data contributing to the detected anomalies should also be specified for localizing the anomalies.
Suppose, for example, that there are two types of monitored data; MIB and syslog, and expressed as feature vectors so that the value of each dimension corresponds to the feature of MIB or syslog data,
such as incoming traffic volume of router-A and the number of appearances of syslog ID 1\footnote[1]{We used the online template extraction by Kimura et al.~\cite{kim2015} for converting syslogs into numeric vectors as discussed in Section~\ref{sec:use_fea}.} in router-B.
Even if an AE detects an anomaly, it does not directly indicate the contributing dimensions to the anomaly. Therefore, we cannot distinguish which feature of which data type contributes to an anomaly, which is required for localizing the anomaly.
One might think the reconstruction errors of AEs, which are the differences between input and reconstructed data calculated for each dimension,
seem to be used for estimating the contributing dimensions.
However, anomalies can affect reconstruction errors of all dimensions due to their fully connected nature, causing misestimations, as we discuss in Section~\ref{sec:eva}.

In this paper, we propose an algorithm to estimate contributing dimensions to anomalies detected by autoencoders
for surveillance of large ICT systems.
With our estimation algorithm, we use sparse optimization for keeping the number of contributing dimensions as low as possible to distinguish the contributing dimensions from others.
Through evaluations with network benchmark data and real measured data,
we argue that our estimation algorithm more accurately specifies contributing dimensions of anomalous data
required for localizing anomalies compared to other naive algorithms including that using the reconstruction error.

We also propose a multimodal AE (MAE) for learning the relationships among cross-domain data more effectively.
The multimodality derived from cross-domain monitoring induces two problems for AEs: one is a combinatorial increase in normal states to be learned, and the other is the difference in learnability among data types.
With our MAE, cross-domain data are integratedly analyzed using a deep AE, which effectively reduces the normal states to be learned and takes into account the difference in learnability by pre-training and weighing anomaly scores.
The evaluation with real measured data showed that our MAE detects anomalies
more accurately than a normal AE.

This paper is organized as follows. We first summarize related work in Section~\ref{sec:related} and explain the principle of AEs in Section~\ref{sec:ae}. The proposed estimation algorithm and MAE are explained in Sections~\ref{sec:spa} and~\ref{sec:mae}, respectively. After evaluating our estimation algorithm in Section~\ref{sec:eva} and discussing a use case of both algorithms with real measured data in Section~\ref{sec:use}, we conclude our paper in Section~\ref{sec:con}.

\section{Related Work}\label{sec:related}
Network anomaly detection has been an important research area; therefore, much work on it has been published. A typical approach is traffic monitoring~\cite{li2006mind, li2006sketch, rin2007}. MIND~\cite{li2006mind} was introduced as a distributed network monitoring infrastructure for detecting anomalous traffic patterns.
Sketches~\cite{li2006sketch} and PCA~\cite{rin2007} were proposed as
anomalous traffic-detection algorithms on the basis of dimensionality reduction.
Performance anomaly detection by system monitoring has also been extensively discussed~\cite{she2009, tan2010}. Shen et al.~\cite{she2009} proposed an algorithm for identifying the symptoms and causes of performance anomalies by analyzing monitored data related to I/O system performance.
Tan et al.~\cite{tan2010} proposed ALERT for raising an anomaly alert in advance by using an anomaly-prediction model according to monitored workload metrics. Other studies focused on syslogs~\cite{qiu2010, kim2015}. SyslogDigest~\cite{qiu2010} and online template extraction~\cite{kim2015} were proposed for extracting network events from syslogs. With these methods, however, monitoring of the data of a single domain is assumed.

Cross-domain monitoring has also been studied~\cite{mah2007, mah2009, nec, ibm}. NICE~\cite{mah2007} is an infrastructure for analyzing pairwise spatio-temporal statistical correlations among cross-domain data that enables anomalous network conditions to be identified. SYNERGY~\cite{mah2009} also detects anomalies by statistically correlating cross-domain data.
Although these systems can not only detect anomalies but also specify their root cause via cross-domain analysis, they are not sufficient for detecting varying and emerging anomalies that appear as collapses of cross-domain relationships.
This is because the former uses rule-based alerting and the latter uses conventional naive anomaly-detection algorithms which do not take into account the relationships among monitored data. NEC Invariant Analyzer~\cite{nec} and IBM Operations Analytics - Predictive Insights~\cite{ibm} have been proposed to detect anomalies on the basis of the analysis of normal relationships among cross-domain data. Although these systems enable varying and emerging anomalies to be detected, linear relationships among data are assumed with both; therefore, they are insufficient for learning complicated correlations among cross-domain data that include syslogs.

AEs~\cite{tho2002, sak2014, ara2016, zho2017} have been attracting much attention for anomaly detection based on nonlinear learning. Thompson et al.~\cite{tho2002} evaluated an AE for detecting the novelty in server utility data. Sakurada and Yairi~\cite{sak2014} discussed anomaly detection using an AE with spacecraft telemetry data, and Araya et al.~\cite{ara2016} introduced an anomaly-detection framework using an AE for detecting abnormal consumption behavior of smart buildings. However, they did not discuss estimating contributing dimensions to the detected anomalies for localizing failures since surveillance of a simplex system is assumed.
Tagawa et al.~\cite{tag2015} proposed a structured denoising AE that enables the analysis of contributing dimensions in accordance with the reconstruction error; however, it requires prior knowledge about the data structure and is not suitable for large and complex ICT systems composed of a huge amount of equipment.
Zhou and Paffenroth~\cite{zho2017} proposed a robust AE that eliminates outliers
from training data as an extension of robust PCA; however, they did not discuss estimating anomalous dimensions in test data.

Multimodality derived from cross-domain monitoring is also problematic for using AEs.
Ikeda et al.~\cite{ike2016} proposed a combination method of multiple AEs trained for each domain, but they ignored cross-domain correlation of data. Although multimodal AE models~\cite{wan2015, jaq2017} have been proposed for cross-domain learning, these models are for encoding multimedia data
or filling in missing data,
and their application to anomaly detection has not been discussed.

\section{Autoencoder}\label{sec:ae}
AEs~\cite{tho2002, sak2014, ara2016, zho2017}, which are artificial neural networks originally used for dimensionality reduction, can be exploited for anomaly detection as follows. By expressing input data as $N$-dimensional vector $\bm{x}^{(1)}$, the values of the $l$-th hidden layer $\bm{x}^{(l)}$ ($l=2,...,L-1$) and $N$-dimensional output layer $\bm{x}^{(L)}$ are calculated using the following recursive formula:
\begin{equation}
\bm{x}^{(l)} = \phi^{(l)}( W^{(l)} \bm{x}^{(l-1)} + \bm{b}^{(l)}), \; l=2,...,L
\end{equation}
where $W^{(l)}$ are the weights between the $(l-1)$th and $l$-th layers, $\bm{b}^{(l)}$ are the biases in $l$-th layer, and $\phi^{(l)}$ is the activation function. From this, we omit subscript (1) of $\bm{x}^{(1)}$ and simply denote it as $\bm{x}$. AEs reconstruct the input data in the output layer, which corresponds to dimensionality reduction in a hidden layer since the size of a hidden layer is set lower than that of the input data. Activation functions perform element-wise transformation and enable AEs to learn nonlinear dimensionality reduction.

In using anomaly detection, by training an AE so that it reconstructs the input data in the output layer when normal data is given, the anomaly score of test data $\bm{x}$ is defined using the following mean squared error (MSE)~\cite{ara2016}:
\begin{equation}\label{eq:mse_test}
MSE(\bm{x}) = \frac{1}{N} \sum_{i=1}^N ( x^{(L)}_{i} - x_{i} )^2.
\end{equation}
Since AEs learn low-dimensional expressions of the normal data in accordance with normal correlations among the dimensions, the MSE becomes higher when abnormal data, in which the normal correlations are collapsed, are input; therefore, it is assumed as the anomaly score of the input data. Training AEs involves optimizing parameters, i.e., $W^{(l)}$ and $\bm{b}^{(l)}$ for $l=2,...,L$. By using training dataset $\bm{x}_1,...,\bm{x}_T$, which contains the data in normal situations, the parameters are optimized so that the mean of the MSE with the training data is minimized by solving the following problem:
\begin{equation}\label{eq:mse_train}
\min_{W^{(l)}, \bm{b}^{(l)}} \frac{1}{T} \frac{1}{N} \sum_{t=1}^T \sum_{i=1}^N ( x^{(L)}_{t, i} - x_{t, i} )^2, \; l=2,...,L,
\end{equation}
where $x^{(L)}_{t, i}$ and $x_{t, i}$ are the $i$th dimensional value of $\bm{x}^{(L)}_t$ and $\bm{x}_t$, respectively.
A representative algorithm for solving problem~(\ref{eq:mse_train}) is stochastic gradient descent (SGD)~\cite{bot2010}, and improvements in SGD have been thoroughly discussed.

\section{Algorithm for estimating contributing dimensions}\label{sec:spa}
Although AEs are promising for use as anomaly-detection algorithms in which labeled data of fault situations cannot be sufficiently obtained, they only calculate the anomaly score as an MSE and do not directly indicate the dimensions contributing to the anomaly. To adopt AEs as anomaly-detection algorithms for surveillance of large ICT systems, however, not only the anomaly score of ICT systems but also the dimensions of the test data that contribute to the anomaly must be distinguished. Let us suppose the use of AEs for network surveillance by monitoring the traffic volume from MIB data. By selecting the traffic volume of each node in a time slot as the value of each dimension of the input data, AEs learn the relationships among the traffic volume of network nodes in a normal situation and detect anomalous behavior as a collapse of the normal relationship. If anomalies are detected, the contributing dimension of the input data judged as abnormal should be clarified to estimate in which node the fault occurs.
For estimating the contributing dimensions of anomalies, our proposed algorithm uses sparse optimization. The following assumption is required with our algorithm.

\textit{Assumption:} The contributing dimensions of an anomaly are a small part of the all dimensions, and if the values of the contributing dimensions are fixed to plausible values in accordance with the learned relationships from the training data, the anomaly score output by an AE will become low.

This assumption is reasonable for the surveillance of large ICT systems in which a massive amount of data is monitored from a number of pieces of equipment and a fault only affects some of the monitored data initially.
According to this assumption, if test data $\bm{\bar{x}}$ are judged as abnormal, we can estimate the contributing dimensions
by finding some of the dimensions that should be amended for decreasing the MSE.

With our algorithm, vector $\bm{\eta}$ is introduced to express how much $\bm{\bar{x}}$ differs from the plausible value for each dimension
and is defined as \textit{contribution degree}, which expresses the contribution of the dimension to the anomaly.
If $x_i$ is higher than a plausible value, $\eta_i$ becomes positive and vise versa.
Decreasing $\eta_i$ from $x_i$ should therefore lower the MSE according to the assumption.
Consequently, the proposed algorithm to determine the contribution degree $\bm{\eta}$ is expressed as the following minimization problem.
\begin{equation}\label{eq:spa}
\min_{\bm{\eta}} MSE(\bar{\bm{x}} - \bm{\eta}) + \lambda ||\bm{\eta}||_{1},
\end{equation}
where the left term expresses minimizing the MSE by moving the input data. The right term is the L1-norm of $\bm{\eta}$, which is responsible for keeping the number of non-zero dimensions in $\bm{\eta}$ as low as possible. This is needed to distinguish contributing dimensions of anomalies, which correspond to non-zero dimensions, from irrelevant dimensions, which correspond to zero dimensions. In this paper, instead of exactly solving Eq.~(\ref{eq:spa}),
we determine $\bm{\eta}$ by iterating the proximal gradient method~\cite{dau2004} with Eq.~(\ref{eq:spa}) until $MSE(\bar{\bm{x}} - \bm{\eta})$ becomes lower than the predefined threshold for computational efficiency.
For obtaining higher sparse $\bm{\eta}$, we execute our algorithm with several $\lambda$.

Estimating the contributing degree can be done for localizing anomalies in large ICT systems as follows.
An anomaly is easily localized if the contributing dimensions simply indicate the location of fault occurrence (for example, each dimension corresponds to the traffic volume in each node in a network and only one dimension is estimated as a contributing dimension).
However, there is still a problem for localization if the contributing dimensions include multiple pieces of information in multiple locations. Clustering is a promising method in such a situation. Since the contribution degree output from the proposed algorithm is a numeric vector, we can cluster the detected anomalies in accordance with the vector of the contribution degree. This will ease the burden of operators for diagnosing anomalies, and once the root cause of the anomalies is identified, the anomalies are ``labeled`` and can be automatically localized when they occur again. This possibility of semi-supervised localizing is detailed in an evaluation with network benchmark data in Subsection~\ref{sec:eva_ben}.

\section{Multimodal autoencoder}\label{sec:mae}
Cross-domain data monitored from ICT systems should be analyzed integratedly for detecting various anomalies that appear as the collapse of cross-domain relationships. However, there are two problems in using AEs for cross-domain data. (i) Although cross-domain relationships should appear in some metrics in data, many metrics have no relationship with those in other types of data. Therefore, the size of ``normal states'' should increase combinationally as the number of monitored metrics increases. However, the volume of training data that can be collected in a real environment is limited and may not be sufficient to learn the normal states. (ii) Learnability of data differs in accordance with the type of data, and the difference should be considered for integrative analysis. If two types of data (one in which the normal states are easy to learn and the other in which they are difficult) are analyzed using an AE, for example, the reconstruction error of the former type will become relatively low and that of the latter type will become relatively high. This difference affects both training and testing, as discussed later; therefore, it should be considered.

\begin{figure}[t]
  \begin{center}
   \includegraphics[width=80mm]{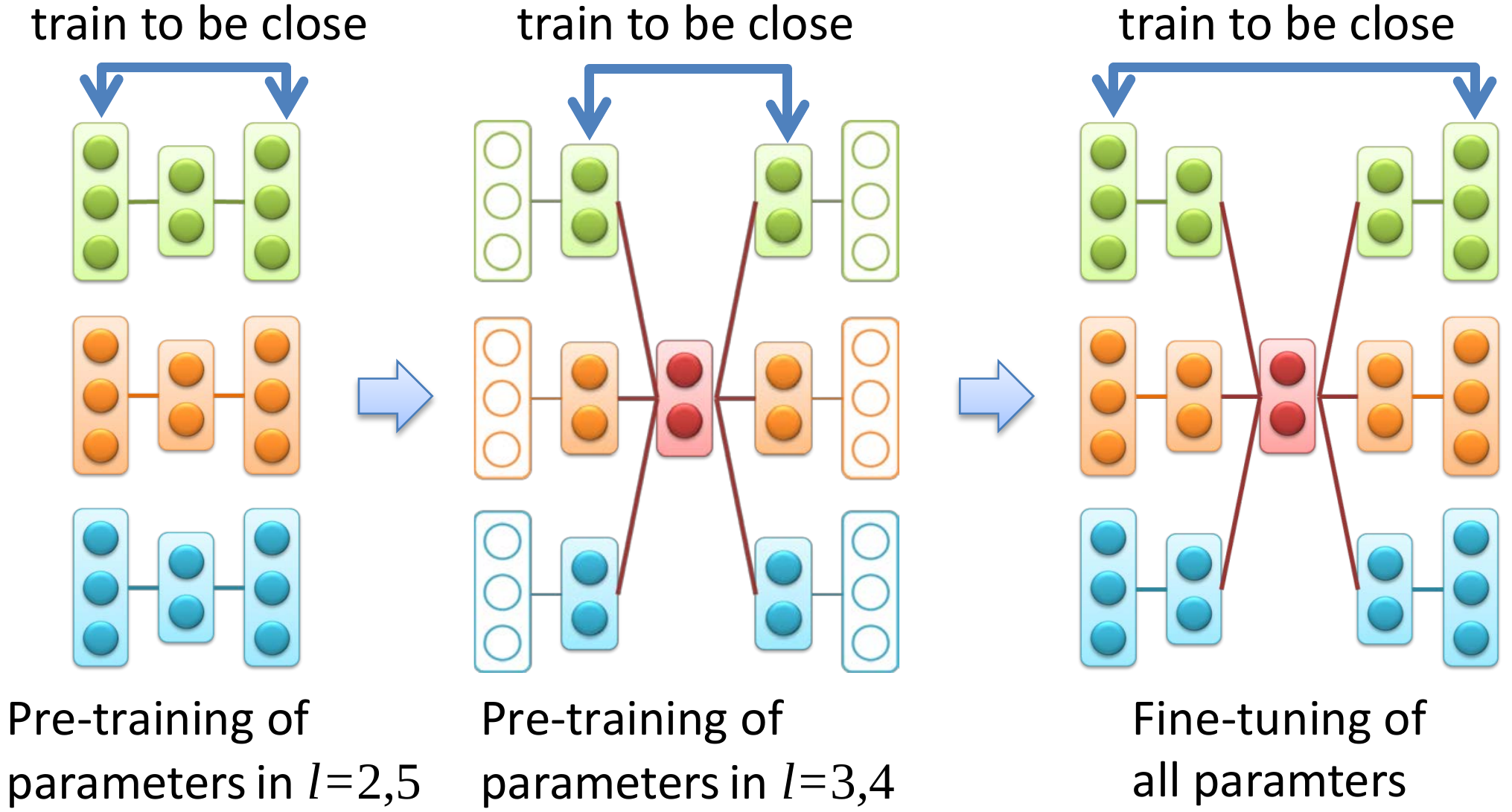}
  \end{center}
  \caption{Training of proposed MAE.}
  \label{fig:mae}
\end{figure}

Taking these problems into account, we propose an MAE for cross-domain data.
We introduce a five-layered model in this paper; however, the number of layers may be increased for more complicated learning.
The first and fifth layers are the input and output layers, respectively, the second and fourth layers learn low-dimensional expressions of each type of training data, and the third layer learns those of all types of training data (see the right figure in Fig.~\ref{fig:mae}). The second and fourth layers are set smaller than each type of data to reduce the dimensions for decreasing the size of normal states to be learned,
which increases according to the combination of the number of the dimensions, as stated in problem (i). By denoting the $k$-th type of input data as $\bm{x}^k$ and $i$-th layer of the $k$-th type of data as $\bm{x}^{k, (i)}$, excluding the third layer for all types of data and denoting it as $\bm{x}^{(3)}$, the values of the second to fifth layers are calculated as follows:
\begin{eqnarray}
\label{eq:mae_l2} \bm{x}^{k, (2)} &=& \phi^{(2)}( W^{k, (2)} \bm{x}^{k} + \bm{b}^{k, (2)}), \; \forall k, \\ 
\label{eq:mae_l3} \bm{x}^{(3)} &=& \phi^{(3)} \biggl( \sum_{k=1}^K W^{k, (3)} \bm{x}^{k, (2)} + \bm{b}^{k, (3)} \biggr), \\
\label{eq:mae_l4} \bm{x}^{k, (4)} &=& \phi^{(4)}( W^{k, (4)} \bm{x}^{(3)} + \bm{b}^{k, (4)}), \; \forall k, \\
\label{eq:mae_l5} \bm{x}^{k, (5)} &=& \phi^{(5)}( W^{k, (5)} \bm{x}^{k, (4)} + \bm{b}^{k, (5)}), \; \forall k, 
\end{eqnarray}
where $K$ is the number of types of data, $\phi^*$ is the activation function, and $W^*$ and $\bm{b}^*$ are the parameters that should be trained by solving the following problem with a training dataset $\bm{x}_{1}^{k},...,\bm{x}_{T}^{k} \; \forall k$:
\begin{equation}\label{eq:mae_mse_train}
\min_{W^*, \bm{b}^*} \frac{1}{T} \sum_{t=1}^T \sum_{k=1}^K \sum_{i=1}^{N_k} \frac{( x^{k, (5)}_{t, i} - x_{t, i}^{k} )^2}{N_k},
\end{equation}
where $N_k$ is the size of the $k$-th type of input data.

Although the proposed MAE is trained to minimize the difference between $\bm{x}^{k, (5)}$ and $\bm{x}^{k}$ similar to the normal AE for all $k$, it requires pre-training and fine-tuning, as shown in Fig.~\ref{fig:mae}, for learning with cross-domain data. In pre-training, $W^{k, (l)},\bm{b}^{k, (l)}, l=2,5, \forall k$ are first trained to reconstruct the training data with Eqs.~(\ref{eq:mae_l2}) and (\ref{eq:mae_l5}) by assuming $\bm{x}^{k, (2)}$ as $\bm{x}^{k, (4)}$ for all $k$. Next, $W^{k, (l)},\bm{b}^{k, (l)}, l=3,4, \forall k$ are trained to reconstruct the dimensionally reduced training data by using Eq.~(\ref{eq:mae_l2}) using already trained $W^{k, (2)},\bm{b}^{k, (2)}, \forall k$ with Eqs.~(\ref{eq:mae_l3}) and (\ref{eq:mae_l4}). In fine-tuning, all parameters are trained using pre-trained parameters as initial values. The importance of pre-training is often discussed in terms of the vanishing gradient problem~\cite{hin2006}. However, it also arises from the differences in learnability among the types of data, as stated in problem (ii). If we try to train the parameters of the proposed MAE at once, reconstruction error $||\bm{x}^{k,(5)} - \bm{x}^{k}||$ tends to be larger than others if the $k$-th type of data are relatively difficult to learn. Such types of data become dominant in minimizing the reconstruction error; therefore, other types of data are not sufficiently reconstructed. Through pre-training, all parameters are first trained to minimize the reconstruction error for each type of data to obtain good initial parameters, and through fine-tuning, the parameters are retrained to minimize the total reconstruction error.
This was numerically evaluated with real measured data, as discussed in Section~\ref{sec:use_eva}.

The difference in learnability should also be considered in calculating the anomaly score of test data. The MSE of the type of data that is relatively difficult to learn tends to be high and vice versa; thus, anomalies in learnable types of data may be overlooked
since the increase in the MSE due to anomalies may be less than the MSE of other types of data in a normal situation.
Therefore, by denoting the $k$-th type of test data as $\bm{x}^{k}$, the anomaly score is calculated as the weighted MSE (wMSE) by taking into account this difference:

\begin{align}\label{eq:mae_mse_test}
wMSE(\bm{x}^{1}, ..., \bm{x}^{K}) &= \sum_{k=1}^{K} w_{k} \frac{\sum_{i=1}^{N_k} ( x^{k, (5)}_{i} - x_{i}^{k} )^2}{N_k}, \\
w_k &= \frac{ 1 / \nu_{k} }{ \sum_{k'=1}^{K} 1 / \nu_{k'} },
\end{align}
where $w_{k}$ is the weight of the wMSE of the $k$-th type of data and $\nu_{k}$ is the mean value of the MSEs of the $k$-th type of data calculated with training data. Since the MSE of the $k$-th type of data tends to be high if $\nu_k$ is high and vice versa, the weight is inversely proportional to $\nu_k$ to offset the difference.

\section{Evaluation}\label{sec:eva}
We evaluated the proposed estimation algorithm with a three-layered AE
using simulated and benchmark data. In all the following evaluations, the input data were normalized to [0,1] by using the minimum and maximum values for each dimension in the training data to align the effects of each feature value to the MSE~\cite{ara2016}.

\subsection{Simulated Data}\label{sec:eva_sim}
We first investigated whether the proposed estimation algorithm can distinguish the contributing dimensions accurately with simulated data. The training data were 1000-dimensional data generated through a simulation as follows:
\begin{equation}
\begin{split}
& x_{i + 10 j} = \begin{cases}
N(1000,200^2) & (j=0) \\
( 1 + 0.1 * j ) * x_i^2 + N(\beta, \gamma^2) & (j=1,...,99)
\end{cases} \\
& \mathrm{for} \; i=1,...,10,
\end{split}
\end{equation}
where $N(\mu, \sigma^2)$ is a random variable of a normal distribution with mean $\mu$ and variance $\sigma^2$. With the data, it was assumed that the surveillance target is composed of ten independent components and there are one random value (such as the number of server requests) and 99 correlation values with noise (such as server load) for each component. For generating test data, we first generated data similar to the training data and randomly chose $i$ from $i = 1,...,10$. After that, we randomly chose $n_f$ values from $x_{i+10j}, j=1,...,99$ and increased (or decreased) the values by $r$-fold, where $r$ was also randomly determined by uniform distribution between [2,10] (or [1/10, 1/2]). This supposed that a fault occurs in a component and affects $n_f$ values correlated with the component. The training and test data were 10,000 and 1 records, respectively. After training the AE, we calculated the MSE of the test data and estimated the contributing dimensions with our estimation algorithm if the MSE exceeded the threshold to examine whether the algorithm can estimate the actual contributing dimensions. A sigmoid function was adopted as an activation function, the size of a hidden layer was set to 10, and batch size and the number of epochs of SGD were both set to 500. The threshold of the MSE was set to $\mu_{mse} + 3 \sigma_{mse}$, where $\mu_{mse}$ and $\sigma_{mse}$ were the mean value and standard deviation of the MSEs with training data, respectively.
Since the MSE exceeded the threshold in all simulations in the following evaluations, we only discuss the estimation accuracy of the contributing dimensions.

\begin{figure*}[!t]
\centering
\subfloat[Outlier degree]{\includegraphics[width=50mm]{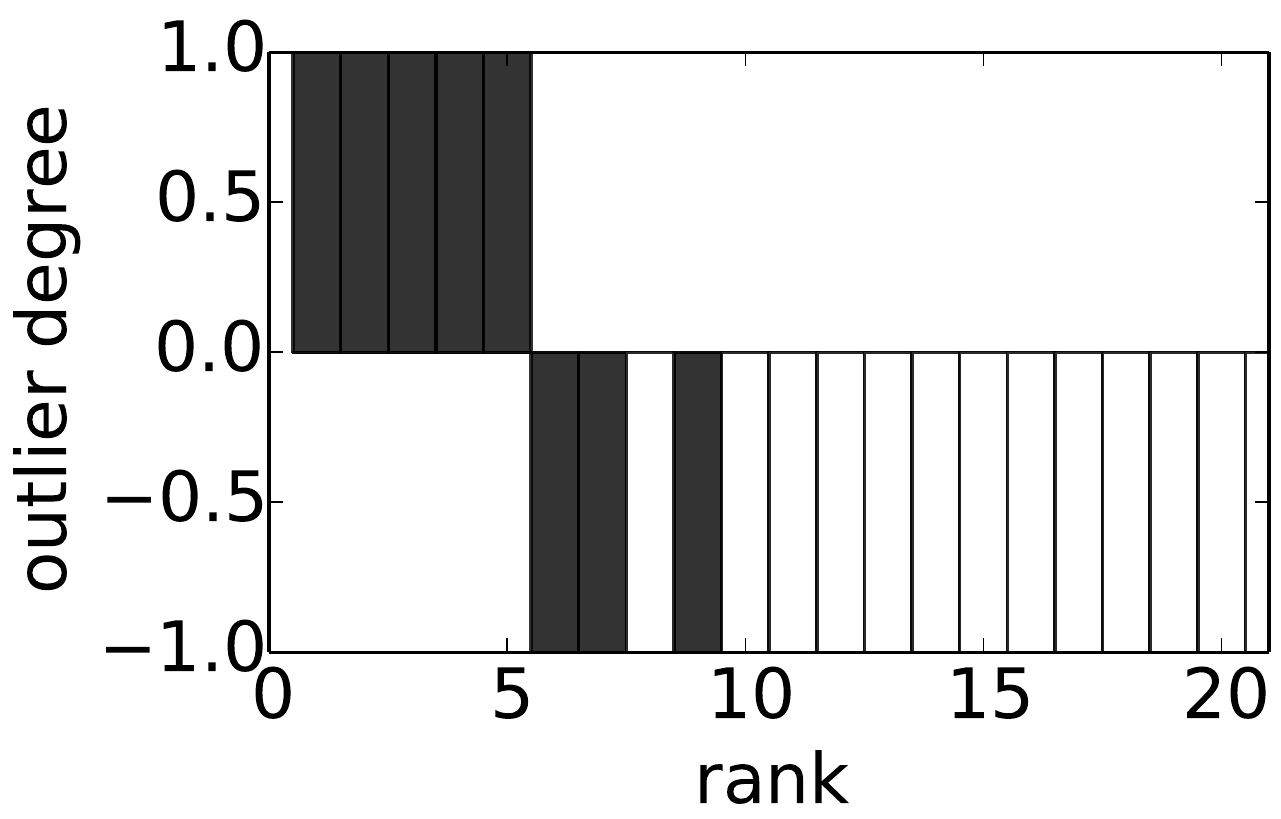}%
\label{fig:graph1_sim_outlier}}
\hfil
\subfloat[Reconstruction error]{\includegraphics[width=50mm]{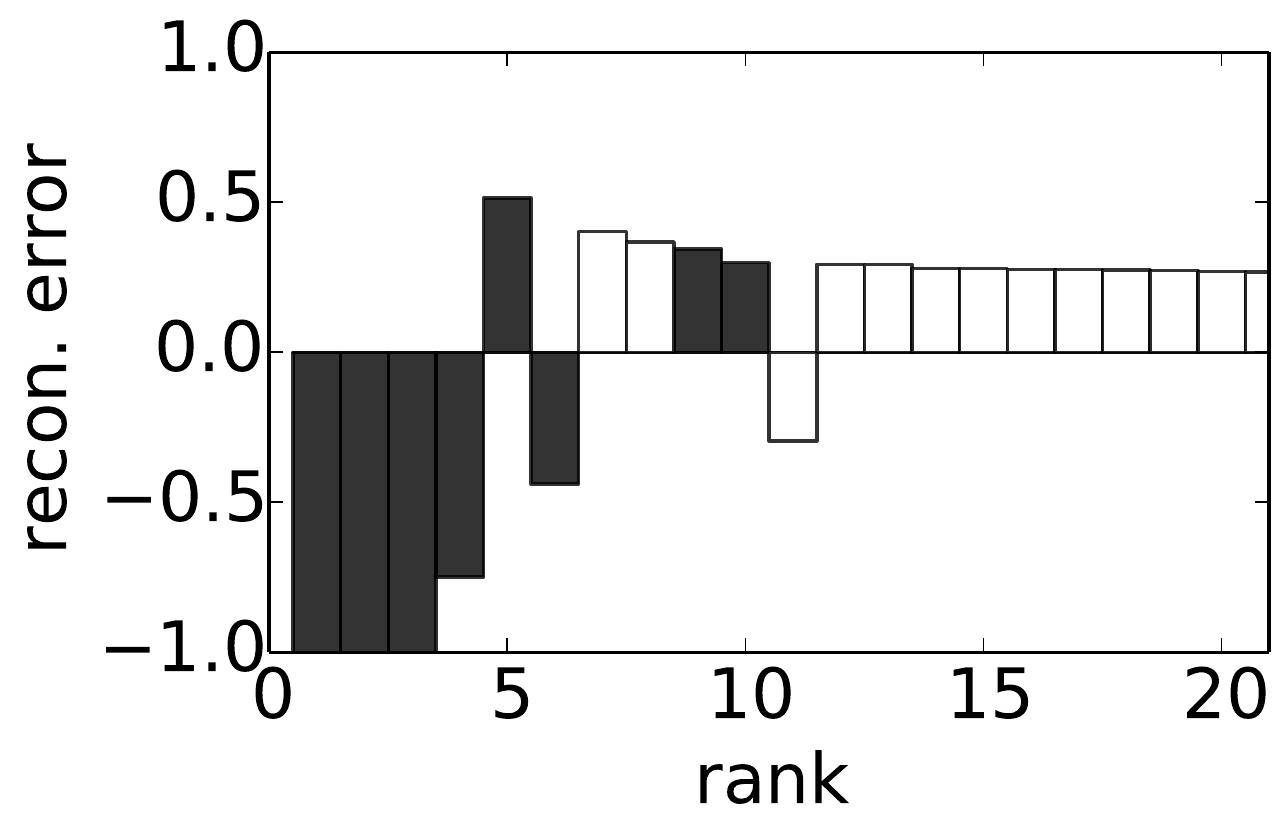}%
\label{fig:graph1_sim_recon}}
\hfil
\subfloat[Contribution degree (w/o L1)]{\includegraphics[width=50mm]{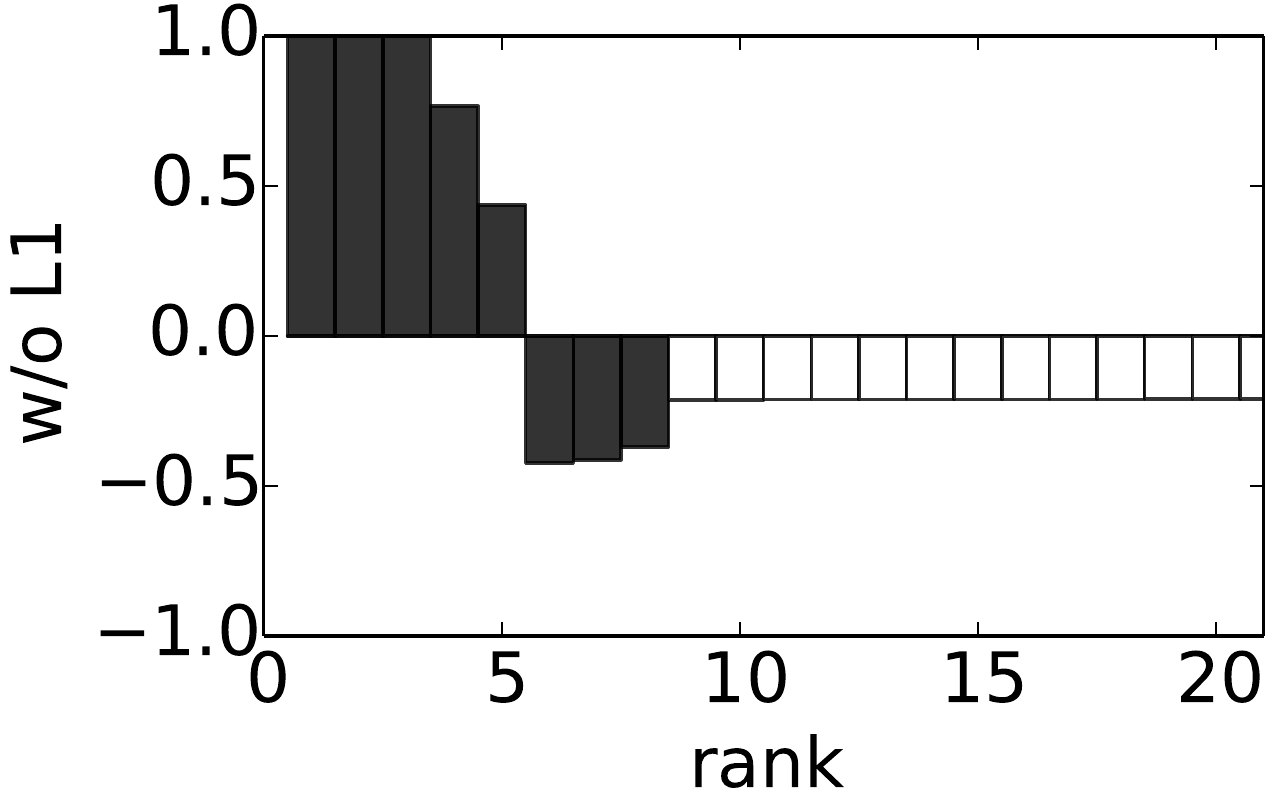}%
\label{fig:graph1_sim_cause_nonspa}}
\hfil
\subfloat[Contribution degree]{\includegraphics[width=50mm]{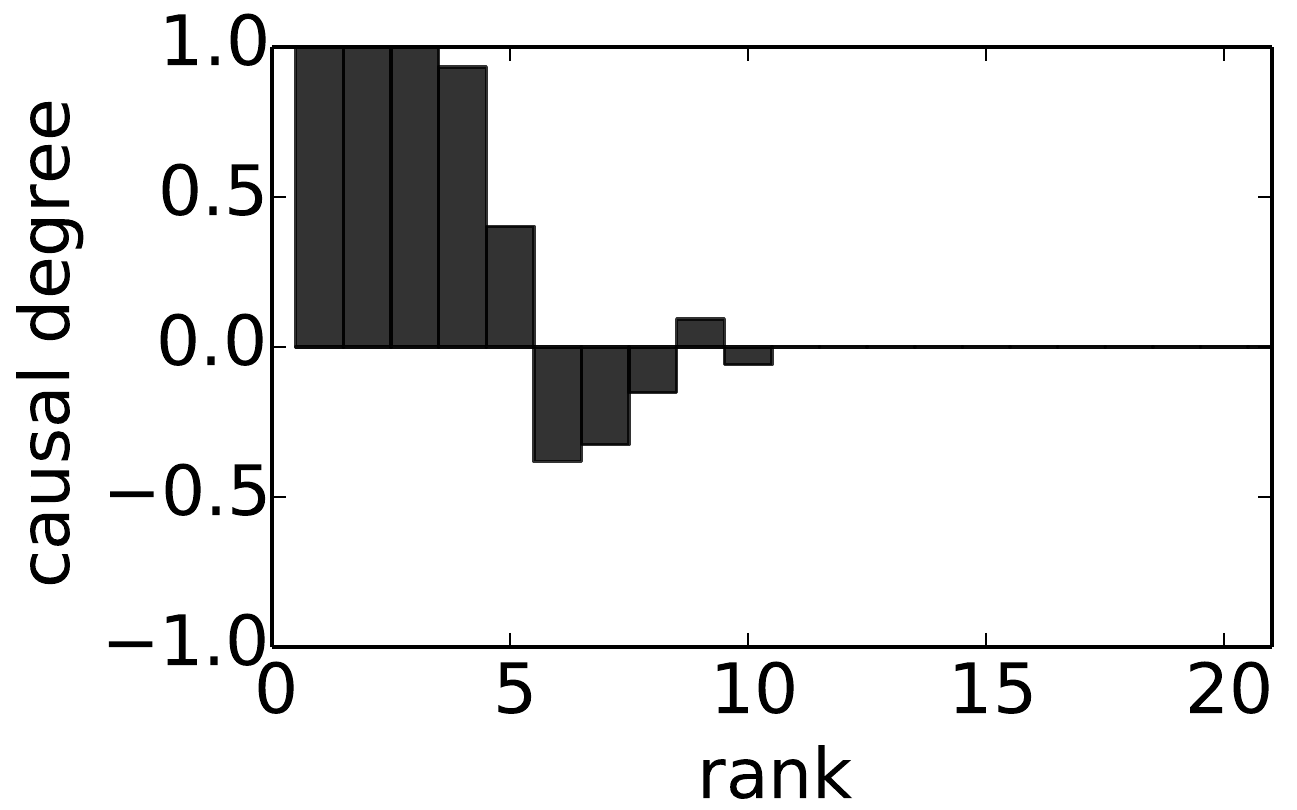}%
\label{fig:graph1_sim_cause}}
\hfil
\caption{Sorted by absolute value}
\label{fig:graph1_sim}
\end{figure*}

We first examine whether the proposed estimation algorithm can distinguish the contributing dimensions from others with $n_f=10$ more accurately than with other naive metrics that may express the contribution of each dimension to the detected anomaly. Figure~\ref{fig:graph1_sim} illustrates the results from a simulation when the dimensions are sorted in accordance with the absolute value, and top 20 values are extracted for each evaluation metric to estimate the contributing dimension, which are \textit{outlier degree}, \textit{reconstruction error}, \textit{contribution degree without sparsity}, and \textit{contribution degree}. The outlier degree of the $i$-th dimension data $x_{test, i}$ is defined using the mean value and standard deviation of the distribution in the training data, $\mu_{train, i}$ and $\sigma_{train, i}$ as $\frac{x_{test, i} - \mu_{train, i}}{\sigma_{train, i}}$. The reconstruction error is the difference between the values of the output and input layers in the AE for each dimension, that is, $x_i^{(L)} - x_i$. The contribution degree without sparsity is a contribution degree calculated without the L1-regularization term. The black and white bars show the contributing dimensions and others, respectively. Note that the vertical axis is limited within [-1:1] to clarify small values, and the top values are actually larger (less) than 1 (-1). In evaluation metrics other than contribution degree, the irrelevant dimensions often have values, and sometimes the value is larger than that of the contributing dimensions. The contribution degree not only reduced the values of irrelevant dimensions to almost 0 but also extracted all 10 contributing dimensions, whereas the other metrics only extracted 8 in the top 20 dimensions.

\begin{figure}[!t]
\centering
\subfloat[$n_f=10$]{\includegraphics[width=50mm]{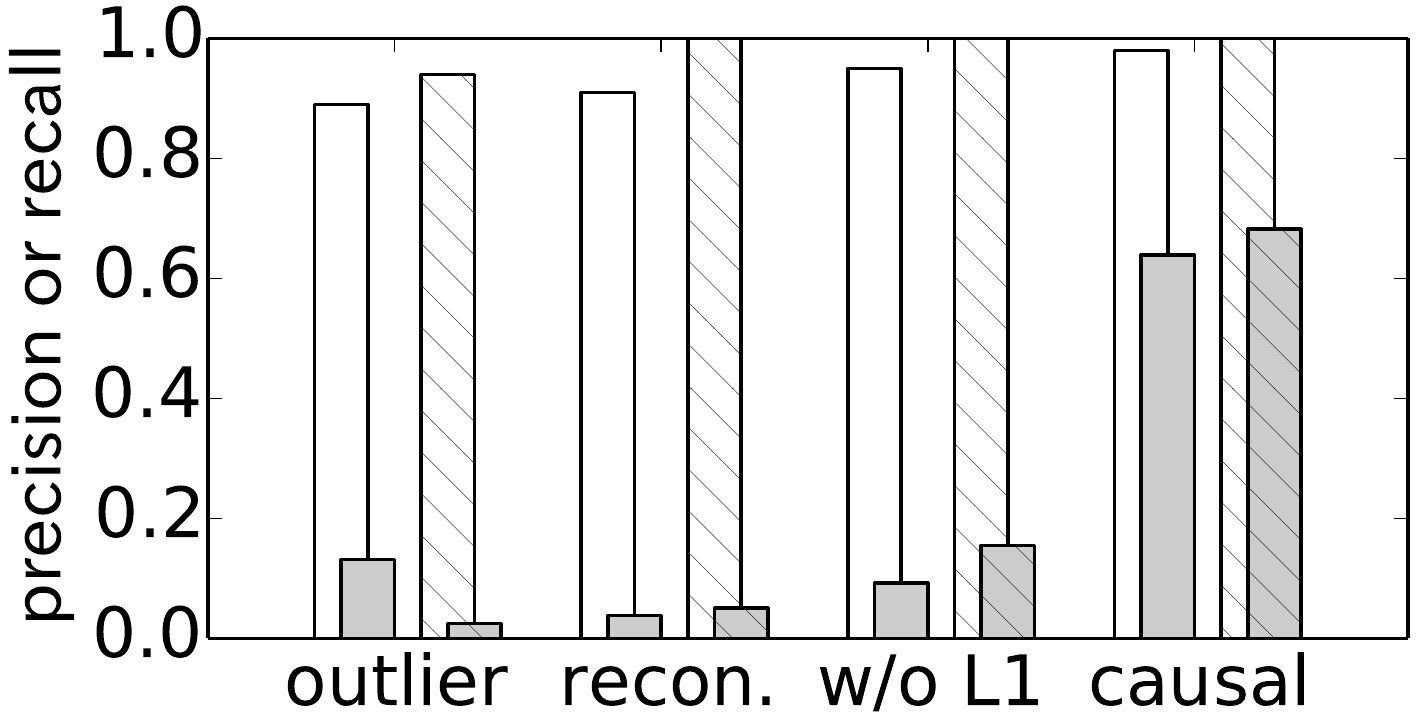}%
\label{fig:graph2_sim_n10}}
\hfil
\subfloat[$n_f=30$]{\includegraphics[width=50mm]{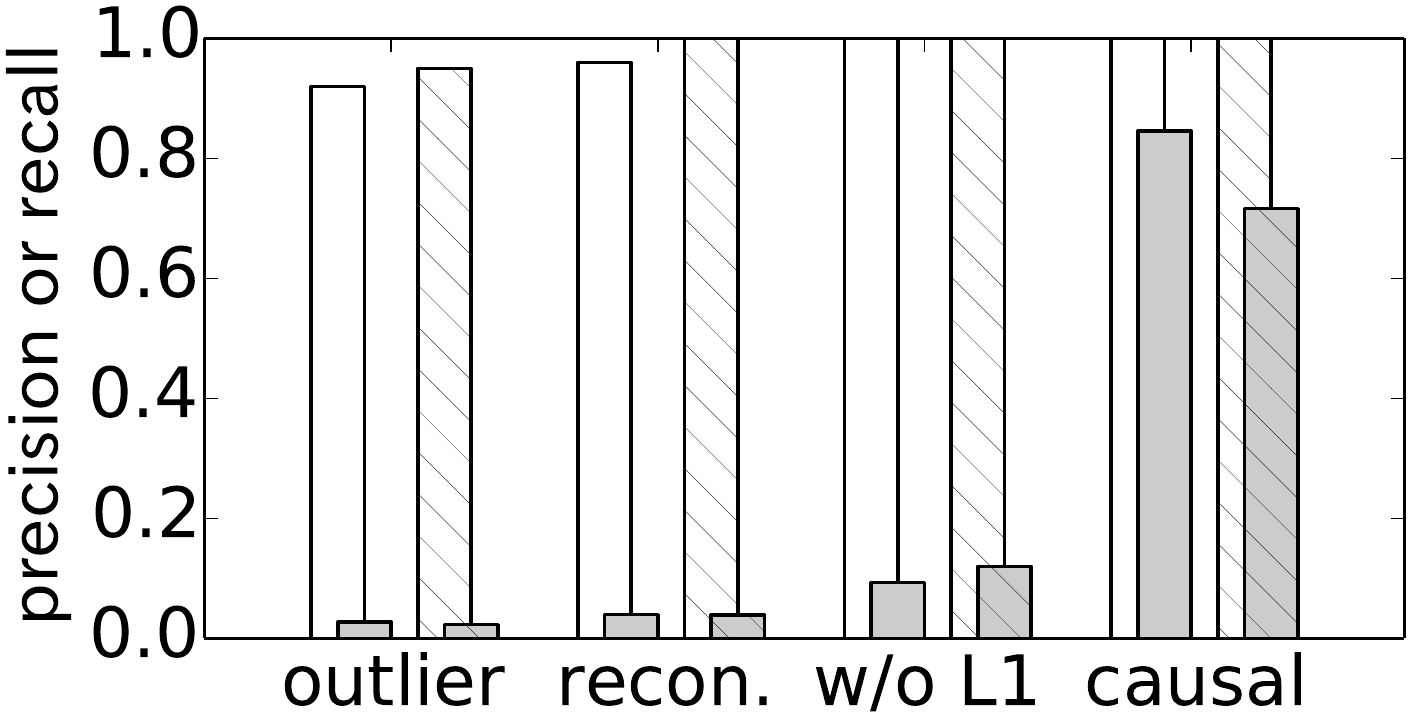}%
\label{fig:graph2_sim_n30}}
\hfil
\caption{Accuracy evaluation. White and grey bars show recall and precision, respectively. Plain bars show results with $(\beta=100, \gamma=50)$, and hatched bars with $(\beta=200, \gamma=50)$.}
\label{fig:graph2_sim}
\end{figure}

We next discuss the estimation accuracy of contributing dimensions by introducing thresholds for each metric.
We defined estimated dimensions as the dimensions that have absolute values of the metric that were larger than the mean absolute value for all dimensions for each evaluation metric. Although the mean absolute value was used as a threshold for comparing the evaluation metrics, the threshold for distinguishing the contributing dimensions is difficult to determine, as explained later. We evaluate recall and precision, which are the ratios of the number of exactly estimated dimensions to that of actual contributing dimensions and predicted dimensions, respectively. The simulation was executed 10 times by randomly generating the training and test data with varying parameters. The mean values of recall and precision are plotted in Fig.~\ref{fig:graph2_sim}. Although recall was relatively high in all evaluation metrics, that is, the estimated dimensions included most actual contributing dimensions, precision was very low except in the contribution degree. This is because all the metrics except for the contribution degree tend to have values in all dimensions, so the predicted dimensions included many irrelevant dimensions. That is, contributing dimensions were not distinguished from others. Even in the contribution degree, however, precision was lower than 1. Since noise was added to the data, the irrelevant dimensions may also diverge from the learned normal correlation; therefore, the sparse optimization in Eq.~(\ref{eq:spa}) sometimes failed to suppress their contribution degree to zero. How to eliminate these dimensions from the contributing dimensions depends on the threshold, and the efficient determination of the threshold should be discussed in future work.

\subsection{Benchmark Data}\label{sec:eva_ben}
Next, we evaluated our estimation algorithm through the NSL-KDD Dataset~\cite{nslkdd}. The dataset consist of 41 feature values about each communication and the communications attribute to five classes. One is normal, and the other four are types of attacks: denial of service (DoS), remote to user (R2L), user to root (U2R), and probing. We used 67,343 normal communications in the training dataset as training data and evaluated the relationship with estimated contributing feature values and its type of communication with 11,850 communications in the test dataset. In the feature extraction, nine discrete feature values were converted into one-hot vectors; consequently, the number of dimensions of feature vectors became 122. We set the size of hidden units of the AE to ten, the activation function as the ReLU function, and the threshold of the MSE as $\mu_{mse} + 3 \sigma_{mse}$, similar to the simulated data discussed in Subsection~\ref{sec:eva_sim}. Although we omit the detailed results of the detection accuracy evaluation since it is not the main scope in this section, the area under receiver operating characteristic (AUROC) curve constructed by plotting the false positive rate (FPR) and true positive rate (TPR) with several threshold values was a maximum of 0.76 with the AE but a maximum of 0.71 with conventional PCA (see details in~\cite{ike2017}). The AE and PCA are also compared in Section~\ref{sec:use}.

\begin{figure}[!t]
\centering
\subfloat[DoS]{\includegraphics[width=50mm]{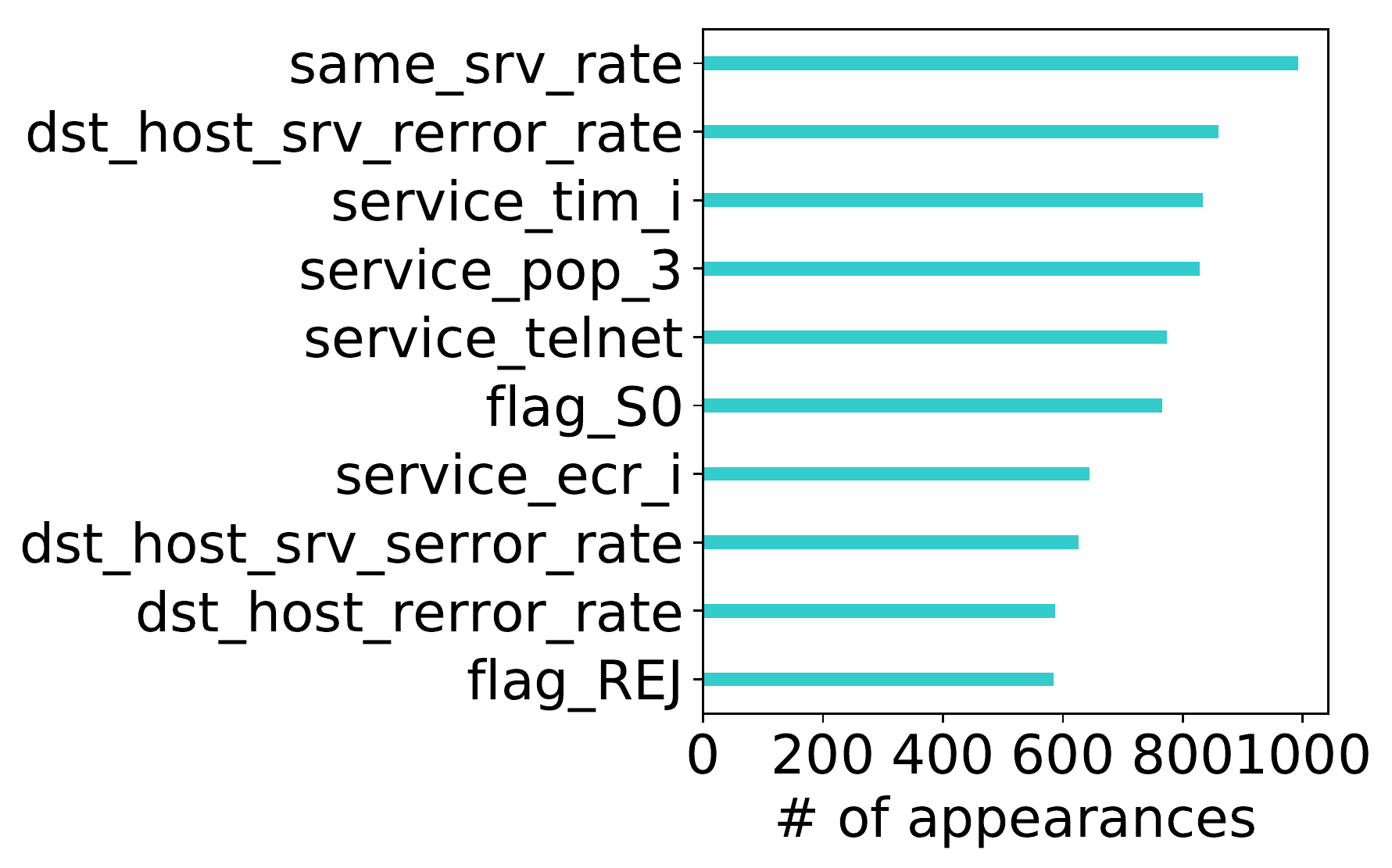}
\label{fig:kdd_top_features_dos}}
\hfil
\subfloat[U2R]{\includegraphics[width=50mm]{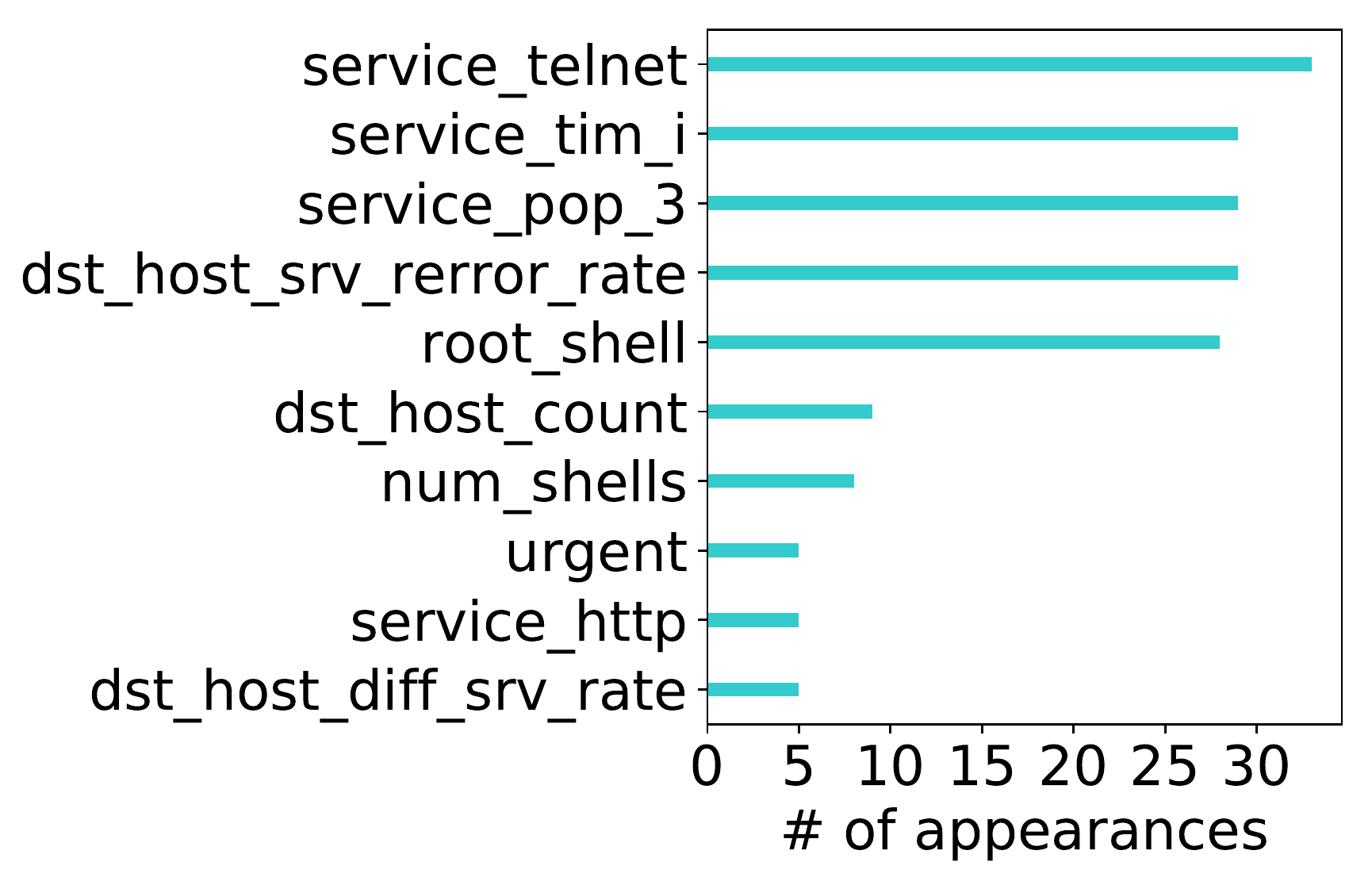}
\label{fig:kdd_top_features_u2r}}
\hfil
\caption{Top features in absolute value of contribution degree}
\label{fig:kdd_top_features}
\end{figure}

\begin{figure}[!t]
\centering
\subfloat[Input data]{\includegraphics[width=50mm]{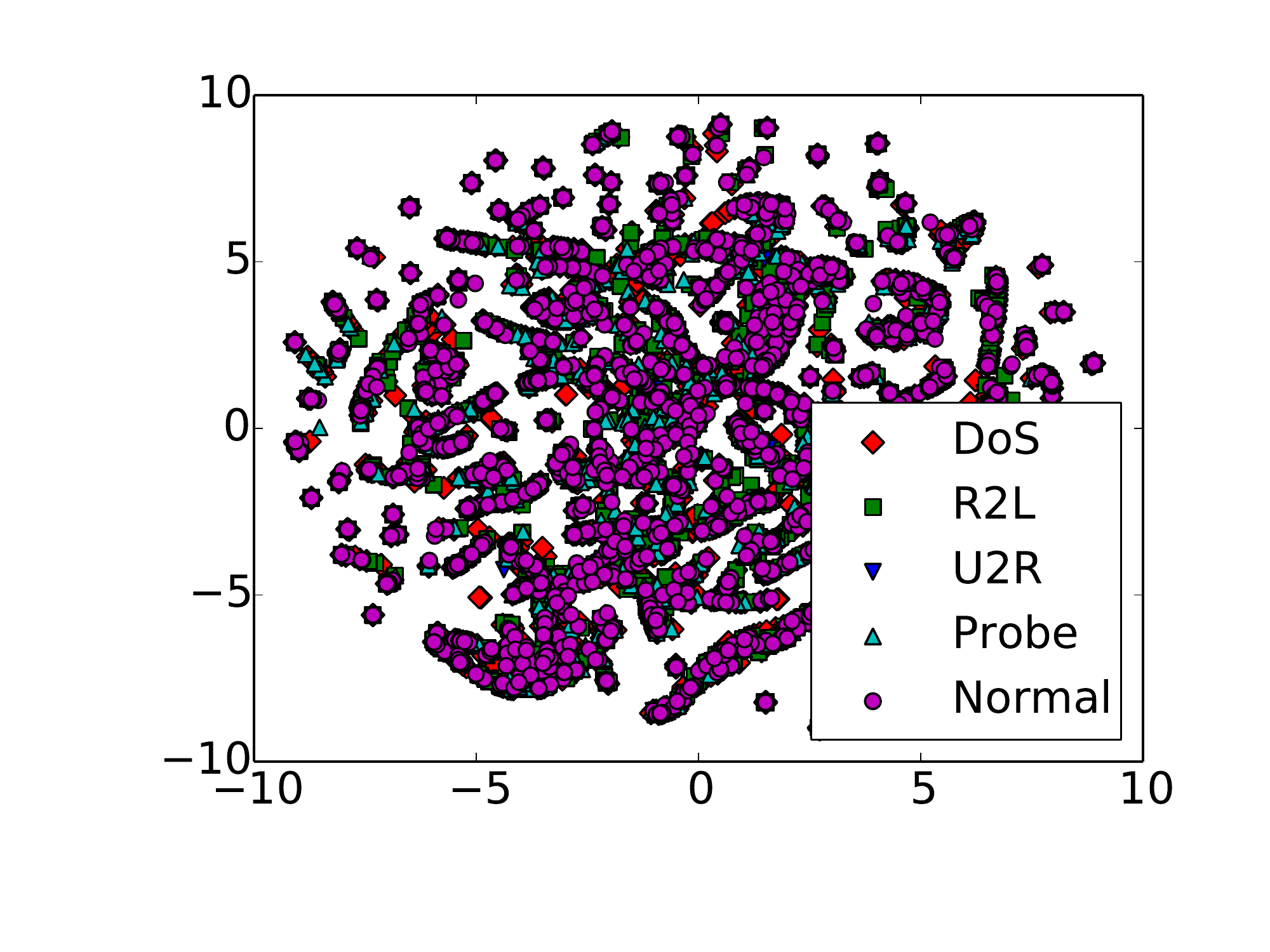}%
\label{fig:kdd_tsne_input}}
\hfil
\subfloat[Contribution degree]{\includegraphics[width=50mm]{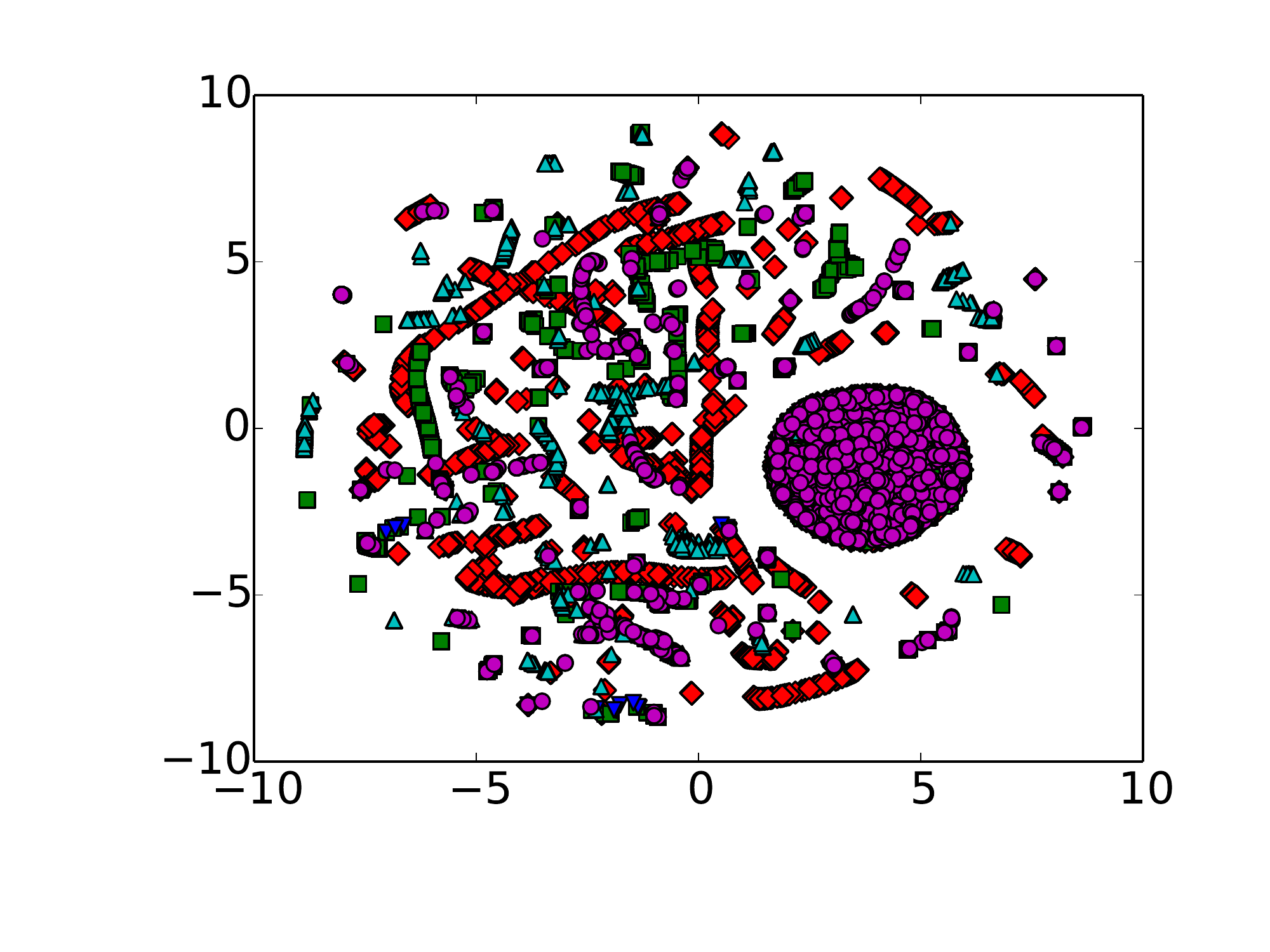}%
\label{fig:kdd_tsne_cause}}
\hfil
\caption{t-SNE plot. Note that a round cluster in the right figure is composed of the data with all the contribution degrees zero, that is, the data not detected as abnormal by the AE.}
\label{fig:kdd_tsne}
\end{figure}

We examined the relationship between feature values with the top absolute value of contribution degree and the type of communications. We counted the number of appearances of features in the top ten absolute values of contribution degrees for individual communications with DoS and U2R, respectively, as shown in Fig.~\ref{fig:kdd_top_features}. The top contributing features in DoS included ``same\_srv\_rate'', which is the ratio of connections to the same service, and ``dst\_host\_srv\_serror\_rate'', which is the ratio of connections from the destination host that have SYN errors. These feature values are intuitively reasonable as the contributing features for detecting DoS attacks including SYN flood. On the other hand, the top contributing features in U2R, which is the attack class of unauthorized access to local root privileges, included ``service\_pop\_3'' and ``root\_shell.'' The former indicates that the service on the destination is POP3, and the latter indicates that a root shell is obtained in the communication. Since POP3 is a representative target of buffer overflow, which is a type of U2R attack, these are also reasonable values of contributing features for detecting U2R communications.

We lastly show a t-distributed stochastic neighbor embedding (t-SNE) plot~\cite{maa2008} in Fig.~\ref{fig:kdd_tsne}. t-SNE is an algorithm for visualizing multidimensional data in accordance with their distance. Therefore, similar datasets are plotted closely, and the distance becomes longer as the two datasets differ. As shown in the figure, although just plotting the test data does not indicate the type of classes, plotting the contribution degree indicates that each cluster seems to be attributed to the same class. This will enable the classification of anomalies by semi-supervised learning with a small number of labels, which should be useful when the labels of failures are not sufficiently obtained.

\section{Use case with real measured data}\label{sec:use}
We show a use case of both the proposed estimation algorithm and MAE by using real measured data and evaluate the effectiveness of using AEs for network anomaly detection while considering multimodality.
The hyperparameters of the anomaly detection models discussed in this section are shown in Table~\ref{tab:testbed_par}.

\begin{table}[t]
\caption{Parameters}
\begin{center}
\begin{tabular}{|p{5cm}|p{5cm}|}
\hline
Hyperparameter & Value \\
\hline
size of principal components of PCA / third layer of AE and MAE & 115 \\
\hline
size of second and fourth layers of AE and MAE & 230 (AE) / 63 for flow, 29 for MIB, and 137 for syslog (MAE) \\
\hline
activation function of AE and MAE & ReLU (second and fourth layers) / None (other layers) \\
\hline
\# of epochs & 200 (AE) / 100 for pre-training and 100 for fine-tuning (MAE) \\
\hline
weight decay & 1E-6 \\
\hline
batchsize & 50 \\
\hline
\end{tabular}
\label{tab:testbed_par}
\end{center}
\end{table}

\subsection{Data Description}
The data are collected from a testbed network consisting of more than 100 nodes over 43 days. The types of data are network-flow data captured at 15 points, MIB data monitored at 36 nodes, and syslog data gathered from 117 nodes. Note that these datasets contain a part of the data in the network since the MIB and syslog data are not gathered from all the nodes and the flow data that did not pass through the capture points cannot be monitored.

Twenty system failures in the testbed were reported as trouble tickets by operators after the fact. Therefore, we examined whether the reported failures can be detected by our MAE with the monitored data. To prepare the training data, we first defined \textit{failure} and \textit{maintenance period}. A failure period is when a reported failure could affect the monitored data. Since the actual period in which the failure affected the monitored data cannot be specified, we defined the period as $\pm 6$ hours from the log timestamp, which is reported as a log sample related to the failure in the trouble ticket, as a margin for eliminating the effect of failures from the training data. A maintenance period is when the testbed undergoes maintenance, which is also reported as a maintenance report with exact start and end time information. The data of the first 28 days were used as training data, from which failure and maintenance periods were excluded. As shown in the evaluation results later, the failure period is not perfect since the actual periods affected by failures are unknown, resulting in insufficient training. How to prepare ``clean'' training data with real monitored data should be discussed in future work. We used all the data of 43 days as test data, including failure and maintenance periods and examine if the reported failures are detected.

\begin{table}[t]
\caption{Feature extraction rule for each data. Each key has the listed values and the feature vector consists of all values for all keys.}
\begin{center}
\small
\begin{tabular}{|c|p{2.3cm}|p{2.1cm}|c|}
\hline
Data & Keys & Values & Vector size \\
\hline
flow &  8 src IP groups, 8 dst IP groups, 10 src ports, 10 dst ports, or 9 protocols & \# of flows, mean or standard deviations of \# of packets, bytes, or durations & 315 \\
\hline
MIB &  36 nodes & mean or standard deviations of \# of incoming or outgoing bytes among interfaces & 144 \\
\hline
syslog & 681 syslog template IDs and 1 extra dimension for newly appeared IDs in test data & \# of appearances & 682 \\
\hline
\end{tabular}
\label{tab:testbed_feat}
\end{center}
\end{table}

\subsection{Feature Extraction}\label{sec:use_fea}
From the monitored data, we extracted feature vectors in accordance with the rules shown in Table~\ref{tab:testbed_feat} for each time bin (1-min. interval).
For the flow data, eight src/dst IP groups, the top ten src/dst ports, and all nine protocols appearing in the training data were used for feature extraction.
The feature vectors were expressed as $\{ v_1 \: \mathrm{of} \: k_1, ..., v_M \: \mathrm{of} \: k_1, v_1 \: \mathrm{of} \: k_2, ..., v_M \: \mathrm{of} \: k_N\}$, where $k_i$ is $i$-th key, such as src IP group, and $v_j$ is the $j$-th value, such as the number of flows. Note that the keys of the syslog data were individual syslog template IDs, which were extracted using online template extraction by Kimura et al.~\cite{kim2015} and appeared in the training data, and all newly appeared IDs in the test data. Therefore, the number of dimensions became that of the IDs observed in the training data plus one as an extra dimension for new IDs in the test data.

\subsection{Evaluation Method}
To evaluate detection accuracy, we defined two evaluation metrics (TPR and FPR). TPR is the ratio of detected failures to all 20 failures. Since the exact time the failure affected the data is unknown, we say that failure could be detected if the MSE exceeded the threshold between $\pm 5$ minutes of the log timestamp at least once. To confirm if failure could be truly detected, we investigated not only whether the MSE exceeded the threshold but also whether the top ten features in absolute value of contribution degree indicated that the failure was the reason for the high MSE via checks by a network operator. To determine the FPR, we defined normal test data as the data of the last 15 days (test period excluding the training period), excluded failure and maintenance periods, and calculated the ratio of the number of time bins in which the MSE became larger than the threshold in the normal test data as the FPR. We set the threshold of the MSE to obtain an FPR of 0.03 and compared the TPR among PCA, the five-layered normal AE, and our MAE. Note that this determination of the threshold is just for comparison under the same conditions, and we cannot determine the threshold in accordance with the FPR of the test period in advance.
Although one can determine the threshold according to the distribution of MSEs with training data,
as shown in Section~\ref{sec:eva_ben}, in advance,
adequate determination of the threshold in practical use should be further discussed in future work.

\begin{table}[t]
\caption{Average MSE of training data for each model}
\begin{center}
\small
\begin{tabular}{|c|c|c|c|}
\hline
Model & flow & mib & syslog \\
\hline
AE for each & 2.04E-3 & 3.47E-4 & 3.29E-5 \\
\hline
MAE & 1.42E-3 & 1.80E-4 & 2.99E-5 \\
\hline
MAE w/o pre-training & 1.60E-3 & 2.30E-4 & 3.24E-5 \\
\hline
AE & 9.37E-4 & 3.31E-4 & 9.42E-5 \\
\hline
\end{tabular}
\label{tab:testbed_mse}
\end{center}
\end{table}

\subsection{Evaluation Results}\label{sec:use_eva}
We first evaluated whether our MAE can sufficiently learn the training data by comparing the average MSEs of training data with the models calculated using Eqs.~(\ref{eq:mse_train}) and (\ref{eq:mae_mse_train}) in Table~\ref{tab:testbed_mse}. ``AE for each'' means that three AEs were trained with each type of data, and ``AE'' means that one AE was trained with all data by merging feature vectors into one vector with a total vector size of 1,141. The number of epochs of SGD for our MAE without pre-training was set to 200 for aligning the number of training times for each parameter.
As shown in the table, the MSE with our MAE was lower than that with the AEs trained for each type of data
since our MAE exploits cross-domain relationships for reconstructing the input data.
Although the MSE of the flow with the AE was less than that with other models,
the MSE of the syslog was higher since the priority of decreasing the MSE of the syslog becomes lower than that of flow, as discussed in Section~\ref{sec:mae}.
The AE that ignores this difference can overlook faults occurring in the syslog as the following discussion shows.

Table~\ref{tab:testbed_result} summarizes the detection results for each failure. Note that there were failures in which the MSE became larger than the threshold, but the relationship between detection and failure was not clear, even though we used our estimation algorithm of contribution dimensions. As shown in the table, 9 out of 20 failures were detected and verified with the AE and PCA, and 10 failures with the MAE.
Our estimation algorithm revealed that these failures actually affected multiple types of monitored data; therefore, they were detected. For example, in the detection of TCP SYN (\#3), the number of flows from a certain src IP group appeared in the top features in the absolute value of contribution degree, that is, the AEs detected the TCP SYN because of the increase in the number of flows in the flow data. In addition, ping check NG (\#14) was detected because of a rapid increase in traffic volume in the MIB data in the node adjacent to that in which ping check NG was reported, and the death of RADIUS server (\#5) was detected by the appearance of corresponding syslogs. To detect these anomalies using a conventional rule-based NMS, we need to manually determine appropriate thresholds for individual monitored metrics, the number of which became 1,141 in this use case. In contrast, AEs only require training data in the normal state and a threshold for MSE to determine if the test data are abnormal.

Some failures were overlooked for several reasons, as shown in Table~\ref{tab:testbed_result}. In particular, \#7 was detected with our MAE but not with the AE or PCA. During failure, there were abnormal syslogs that correlated with the failure; therefore, our MSE was affected. With the AE and PCA, however, the fluctuation in the MSE was not noticeable since the MSE of the syslog was lower than that of other types of data. Our MAE could detect the abnormality of syslogs since
it emphasizes the fluctuation by sufficiently learning the normal situation of syslogs via pre-training
and calculates a wMSE to offset the difference in the natural MSE among the types of data.

Although the failures in this use case did not appear as collapses of relationships among cross-domain data, which is the main target with our MAE, such failures should become significant since network virtualization plays a major role, as Govindan et al. discussed~\cite{gov2016}. The efficacy of our MAE for such situations should also be examined in future work.

\begin{table*}[t]
\caption{Results of detection and verification from contribution degrees by network operator. \checkmark \checkmark indicates that the failure is detected and verified, \checkmark that the failure is detected but not verified, and $\times$ that MSE does not reach the threshold.}
\begin{center}
\small
\scalebox{0.8}{
\begin{tabular}{|c|c|c|c|c|c|}
\hline
Failure \# & Description & MAE & Normal AE & PCA & Reason for overlooking \\
\hline
\#1 & dpkg locked & $\times$ & $\times$ & $\times$ & not reacted \\
\hline
\rowcolor[gray]{0.85} \#2 & Disconnection to SNMP master agent & $\times$ & $\times$ & $\times$ & training data included error logs \\
\hline
\#3 & TCP SYN & \checkmark \checkmark & \checkmark \checkmark & \checkmark \checkmark & - \\
\hline
\rowcolor[gray]{0.85} \#4 & SYN Flood & \checkmark \checkmark & \checkmark \checkmark & \checkmark \checkmark & - \\
\hline
\#5 & No response from RADIUS server & \checkmark \checkmark & \checkmark \checkmark & \checkmark \checkmark & - \\
\hline
\rowcolor[gray]{0.85} \#6 & License expired & \checkmark & \checkmark  & \checkmark & - \\
\hline
\#7 & Change of SAP port state 1 & \checkmark \checkmark & $\times$ & $\times$ & reacted but did not reach threshold \\
\hline
\rowcolor[gray]{0.85} \#8 & Detected IP Private & $\times$ & $\times$ & $\times$ & not reacted \\
\hline
\#9 & Virus alert 1 & $\times$ & $\times$ & $\times$ & not reacted \\
\hline
\rowcolor[gray]{0.85} \#10 & Change of SAP port state 2& \checkmark & \checkmark & \checkmark & - \\
\hline
\#11 & Failed to open /dev/smb & \checkmark \checkmark & \checkmark \checkmark & \checkmark \checkmark & - \\
\hline
\rowcolor[gray]{0.85} \#12 & Unreported SDP update and interface down & \checkmark \checkmark & \checkmark \checkmark & \checkmark \checkmark & - \\
\hline
\#13 & ICMPv6 checksum failed & \checkmark & \checkmark & $\times$ & not reacted \\
\hline
\rowcolor[gray]{0.85} \#14 & ping check NG 1 & \checkmark \checkmark & \checkmark \checkmark & \checkmark \checkmark & - \\
\hline
\#15 & Virus alert 2 & \checkmark & \checkmark & \checkmark & - \\
\hline
\rowcolor[gray]{0.85} \#16 & ping check NG 2 & \checkmark & \checkmark & \checkmark &  \\
\hline
\#17 & Softwares are not updated & \checkmark & \checkmark & \checkmark & - \\
\hline
\rowcolor[gray]{0.85} \#18 & STP state change & \checkmark \checkmark & \checkmark \checkmark & \checkmark \checkmark & - \\
\hline
\#19 & Linux kernel error & \checkmark \checkmark & \checkmark \checkmark & \checkmark \checkmark & -  \\
\hline
\rowcolor[gray]{0.85} \#20 & HA cluster bug & \checkmark \checkmark & \checkmark \checkmark & \checkmark \checkmark & - \\
\hline
\multicolumn{2}{|c|}{TPR of detected / detected and verified failures} & 0.8 / 0.5 & 0.75 / 0.45 & 0.7 / 0.45 & - \\
\hline
\end{tabular}
}
\label{tab:testbed_result}
\end{center}
\end{table*}

\section{Conclusion}\label{sec:con}
We proposed two algorithms for surveillance of ICT systems. One estimates contributing dimensions to anomalies detected using AEs to localize the anomalies in large ICT systems, and the other is an MAE that enables relationships among cross-domain data to be learned. We evaluated the two algorithms with several types of data including real measured data and validated the effectiveness of our estimation algorithm for localizing the detected anomalies and improvement in accuracy of our MAE for detecting anomalies.

Although the contributing dimensions were estimated with our algorithm, there are still problems to specify the root cause of anomalies. If the distinguished contributing features do not directly indicate the root cause, two approaches for specification can be considered. One is cooperating with conventional root cause analysis functions, and the other is semi-supervised labeling by clustering test data in accordance with the contribution degree. These should be discussed in future work.



%

%
%
%
%


\bibliographystyle{ieicetr}
\bibliography{nwai}



\end{document}